\documentclass{article}

\usepackage{microtype}
\usepackage{graphicx}
\usepackage{subfigure}
\usepackage[ruled,vlined]{algorithm2e}
\usepackage{booktabs} 

\usepackage{hyperref}

\usepackage{multirow} 

\usepackage[accepted]{icml2024}

\usepackage{amsmath}
\usepackage{amssymb}
\usepackage{mathtools}
\usepackage{amsthm}

\usepackage{colortbl, xcolor}
\definecolor{grey}{rgb}{0.1,0.1,0.1}

\usepackage[capitalize,noabbrev]{cleveref}

\theoremstyle{plain}

\theoremstyle{definition}

\theoremstyle{remark}

\usepackage[textsize=tiny]{todonotes}

\icmltitlerunning{Fair Text-to-Image Diffusion via Fair Mapping}

\begin{document}

\twocolumn[
\icmltitle{Fair Text-to-Image Diffusion via Fair Mapping}



\icmlsetsymbol{equal}{*}

\begin{icmlauthorlist}
\icmlauthor{Jia Li}{equal,1,2,3}
\icmlauthor{Lijie Hu}{equal,1,2,6}
\icmlauthor{Jingfeng Zhang}{4}
\icmlauthor{Tianhang Zheng}{5}
\icmlauthor{Hua Zhang}{3}
\icmlauthor{Di Wang}{1,2,6}
\end{icmlauthorlist}

\icmlaffiliation{1}{Provable Responsible AI and Data Analytics (PRADA) Lab}
\icmlaffiliation{2}{King Abdullah University of Science and Technology}
\icmlaffiliation{3}{Chinese Academy of Sciences}
\icmlaffiliation{4}{University of Auckland}
\icmlaffiliation{5}{University of Missouri}
\icmlaffiliation{6}{SDAIA-KAUST}

\icmlcorrespondingauthor{Di Wang}{di.wang@kaust.edu.sa}

\icmlkeywords{Machine Learning, ICML}

\vskip 0.3in
]



\printAffiliationsAndNotice{\icmlEqualContribution} 

\begin{abstract}
In this paper, we address the limitations of existing text-to-image diffusion models in generating demographically fair results when given human-related descriptions. These models often struggle to disentangle the target language context from sociocultural biases, resulting in biased image generation. To overcome this challenge, we propose Fair Mapping, a flexible, model-agnostic, and lightweight approach that modifies a pre-trained text-to-image diffusion model by controlling the prompt to achieve fair image generation. One key advantage of our approach is its high efficiency. It only requires updating an additional linear network with few parameters at a low computational cost. By developing a linear network that maps conditioning embeddings into a debiased space, we enable the generation of relatively balanced demographic results based on the specified text condition. With comprehensive experiments on face image generation, we show that our method significantly improves image generation fairness with almost the same image quality compared to conventional diffusion models when prompted with descriptions related to humans. By effectively addressing the issue of implicit language bias, our method produces more fair and diverse image outputs.
\end{abstract}

\begin{figure}[ht]
\vskip 0.2in
\begin{center}
\centerline{\includegraphics[width=\columnwidth]{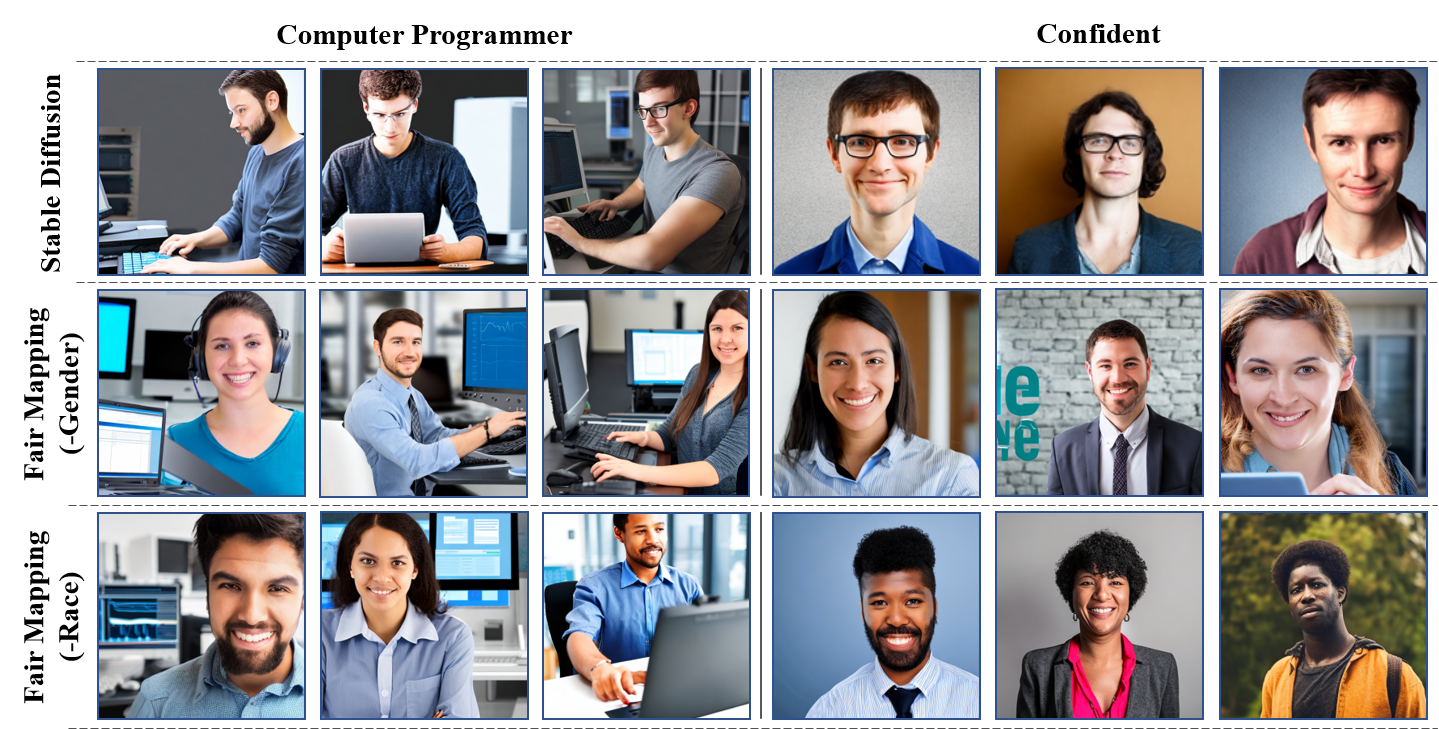}}
\caption{\textbf{Fair Mapping (our method) can balance demographic visual images in text-to-image diffusion models.} Fair Mapping minimally adjusts parameters during training to eliminate demographic biases in pre-trained text-to-image models, resulting in more equitable image generation. Here, Stable Diffusion (top row) runs the risk of lacking diversity in its output, e.g., only male-appearing persons generation as \textit{computer programmer} and \textit{confident}. In contrast, Fair Mapping (with different sensitive attributes) allows the creation of more equitable and unbiased images.}
\label{fig:bg1}
\end{center}
\vskip -0.2in
\end{figure}
\vspace{-0.3in}

\section{Introduction}
\label{submission}

Text-to-image diffusion models \cite{ddpm,ddim} have achieved remarkable performance in various applications \cite{diffusion1_beatgan, textdiffusion7-attactkdiffusion, diffuion-video1, diffusion-video2, diffusion-3d1, diffusion-3d1-humanmotion} by involving incorporation of textual conditional elements with language models \cite{diffusion-c-free}. 
With the increasing popularity among the public, the generation of diverse images \cite{diffusion8-guided_robustness} about human-related description becomes crucial, yet it remains a challenging task \cite{diffusion-bias,safediffusion,text-guided-bias2}. For example, when prompted to ``An image of a computer programmer'' and ``An image of a confident person'' to Stable Diffusion, as exemplified in Figure \ref{fig:bg1}, it is obvious that the majority of results are predominantly white male, even if there are no specified gender or race descriptions in textual condition. 

This phenomenon demonstrates that text-to-image diffusion models learned implicit correlations between demographic visual representations and textual descriptions during training. Real-world data is often biased and incomplete, reflecting inherent stereotypes in human perceptions \cite{data-realworld,text-guided-bias2}. When training on a dataset that includes pairs of images and texts \cite{dataset-laion5b,dataset-laion400}, text-to-image diffusion models face challenges in disentangling sensitive attributes such as gender or race, from specific prompts, which potentially introduce during denoising and lead to societal biases in generation.

Consequently, the pressing question remains: \textit{How can unbiased inferences be made in the face of training data that is inherently biased?} Nowadays, \citet{fair_editing_time,fair_editing_unified} highlight text-to-image models implicitly assume sensitive attributes from the language model, where textual condition embeddings demonstrate an inclination towards specific demographic groups.
As a result, some work \cite{fair_shen2023finetuning,fair_finetuning_fraser2023friendly} takes fine-tune methods to control the distribution of generated images. However, updating the parameters of the diffusion model with expensive data collection potentially results in knowledge forgetting, ultimately detrimental to stable generative ability. \cite{fairdiffusion} allows human instruction, enabling to guide output based on desired criteria. These post-processing methods require significant computational resources and time, which is not efficient enough without a robust framework.

In this paper, we propose a novel post-processing, model-agnostic, and lightweight method namely {\bf Fair Mapping}. Briefly speaking, there are two additional components in Fair Mapping compared to vanilla diffusion models: The first one is a linear mapping network which is strategically designed to rectify the implicit bias in the representation space of text encoder in text-to-image diffusion models. It addresses the disentanglement of the target language context from implicit language biases by introducing a fair penalty mechanism. This mechanism fosters a harmonious representation of sensitive information within word embeddings via a flexible linear network with only a modest addition of new parameters. At the inference stage, Fair Mapping introduces a detector, which aims to detect input containing potential biased content for a robust generation. 
To summarize, our contributions are three-fold.

\begin{itemize}
    \item \textit{Analysis of Language Bias and Proposed Fairness Evaluation Metric:} We first quantitatively explain the bias issue in generated images caused by the textual condition within text-guided diffusion models, providing insights into the contributing dynamics. We introduce evaluation metrics designed to assess the language bias and diffusion bias of diffusion models in generating text-guided images. These metrics provide a systematic and objective measure for quantifying the reduction of bias in the generative process, enabling a more precise evaluation of fairness outcomes. 
    
    \item \textit{Innovative Fair Mapping Module:} To mitigate language bias, 
    we develop a model-agnostic and lightweight method namely Fair Mapping. Generally speaking, Fair Mapping introduces a linear network before the Unet structure, which optimizes minimal extra parameters for training, enabling seamless integration into any text-to-image generative models while keeping their parameters unchanged. At the inference stage, there is an additional detector compared to conventional diffusion models, which aims to detect whether we need to pass the input text prompt given by users through the linear network for debias.
    
    \item \textit{Comprehensive Experimental Evaluations:}
    Finally, we conduct comprehensive experiments of our methods in terms of the fairness and quality of our generated images. Specifically, our experiments demonstrate that our method improves performance and outperforms several text-to-image diffusion models for fairness based on human descriptions. We show that the image quality of generated content is very close to and even better than that of the base diffusion models.
\end{itemize}

\section{Related Work}\label{sec:related_work}
\noindent {\bf Text-guided Diffusion Models:} Text-guided diffusion models merge textual descriptions with visual content to create high-resolution, realistic images that align with the semantic guidance provided by the accompanying text prompts \citep{textdiffusion2, textdiffusion3, textdiffusion1, textdiffusion4-desentext, textdiffusion5-SpaText, textdiffusion6-eDiff-I, textdiffusion7-ernie,textdiffusion8-attention}. However, this fusion of modalities also brings to the forefront issues related to bias and fairness \citep{text-guided-bias1,text-guided-bias2}, which have prompted extensive research efforts to ensure that the generated outputs do not perpetuate societal inequalities or reinforce existing biases. In this paper, we explore these challenges and the state-of-the-art solutions aimed at enhancing the fairness and equity of text-guided diffusion models.

\noindent {\bf Bias in Diffusion Models:} While large datasets are commonly used in data-driven generative models, they often contain social and cultural biases \citep{dataset-multimoda, dataset-laion5b, dataset-laion400}. Previous efforts have addressed this challenge by optimizing model parameters after training \cite{fair_shen2023finetuning,fair_finetuning_fraser2023friendly}, while \cite{fair_rs_jiangyue} accomplish distributional control by updating the latent code. \citet{fairdiffusion} introduces a post-processing mechanism based on human instructions. \citet{fair_language_chuang2023debiasing} remove biased directions in text embeddings to mitigate bias in vision-language. Similarly, \citet{fair_editing_time,fair_editing_unified} update the model's cross-attention layers to achieve concept editing in certain text for debiasing. In our work, we also seek to bridge this gap by addressing language biases in semantic representation space, thereby contributing to a more comprehensive understanding and mitigation of biases in generative data by employing an end-to-end framework.

\noindent {\bf Bias in Language Models:} The Transformer structure of language models is capable of storing and encoding knowledge, including societal biases reflected in the training data. This capability is also extended to text-to-image diffusion models through the incorporation of attention layers \cite{fair_language_meng2022mass,fair_language_arad2023refact,fair_language_berg-etal-2022-prompt}. Ensuring fairness in these models has been extensively studied and validated, especially in the context of large-scale models \citep{lms-bias1, lms-bias2, lms-bias3, lms-bias4,lms-bias5,lms-bias6,lms-bias7}. Efforts have been made to address these biases, with approaches introduced by \citet{fairclip, fairclip2} aiming to mitigate the impact of bias. 
In our work, we explore the intersection of bias mitigation efforts in language models, which is a critical juncture in the pursuit of fairness and ethics in artificial intelligence.

\vspace{-0.1in}
\section{Language Bias in Text-to-Image Diffusion Models}\label{sec:language_bias}
We first give some notations that will be used throughout the paper.

{\noindent \bf Notations.} Consider a keyword set $C$ such as a set of different occupations. For each keyword $c$, such as ``doctor'', it has a set $A$ of possible sensitive attributes such as ``male'' or ``female''. \footnote{Here we suppose the attribute set is the same for all keywords for convenience.} For language bias, we denote $prompt(a, c)$ for each $a\in A, c\in C$ as a prompt in a uniform and specific format. For example,  $prompt(\text{male},\text{doctor})=$ ``an image of a male doctor''. We also denote $prompt(\text{` '}, c)$ as the prompt where there is no sensitive attribute, such as $prompt(\text{` '}, \text{doctor})=$ ``an image of a doctor''. Given the text encoder for textual conditioning in a text-to-image diffusion model, we extract text representations $f$ and $\{f_j\}_{j=1}^{|A|}$ from $prompt(\text{` '}, c)$ and $\{prompt(a_j, c)\}_{a_j\in A}$ respectively. These representations are essential in conventional text-guided diffusion models for generating coherent and contextually relevant text samples.

Before showing our method for mitigating language bias, it is necessary to address the following fundamental question: \textit{Whether there indeed exists implicit language bias even if there is no explicit sensitive attribute in the textual information of prompt?} 
Based on the above notation, we propose the bias metric for the input prompts and the generated outputs of text-to-image diffusion models.

1) {\bf Diffusion Bias:} We propose a novel evaluation metric based on group fairness to robustly assess fairness in the generative results of diffusion models across diverse groups. This metric captures variations in generated outcomes among demographic attributes \citep{metric-odds}, such as gender and race, and quantifies fairness by evaluating equilibrium. Our study adopts a highly specific and constrained definition of fairness in the evaluation process. 
A diffusion model is absolutely fair if it satisfies that for any keyword $c_k\in C$
\begin{equation}
\small
    P(s=a_i|c=c_k) = P(s=a_j|c=c_k), \text{for all } a_i, a_j \in A, \label{eo}
\end{equation}
where $s$ is the random variable of the sensitive attribute for the output of the diffusion model, $c$ is the random variable of the keyword in conditional textual information, $P(s=a_i|c=c_k)$ represents the probability of the sensitive attribute $s$ of generative images expressing $a_i$ given a specific conditional prompt with keyword $c_k$. We define our diffusion bias evaluation criteria towards attribute $a_i$ for $c_k$ as follows:
\begin{multline}
\small 
     DBias_{a_i}(c_k)=P\left(s=a_i \mid c=c_k\right) \\ - \frac{1}{\mid A \mid}  \sum_{a_j\in A} P\left( s=a_j \mid c=c_k\right),
\end{multline}
   

Thus, based on \eqref{eo}, for a keyword $c_k$, our metric on the diffusion bias is defined as follows:
\begin{equation}
\small
\text{BiasScore}(c_k)= \sqrt{\frac{1}{\mid A \mid} \sum_{a_i\in A } \left( DBias_{a_i}(c_k) \right)^2}. 
\end{equation}
Thus, for a dataset $C$ that contains keywords, our fair evaluation metric on the diffusion bias is $\frac{1}{|C|} \sum_{c_k \in C} \text{BiasScore}(c_k)$.
A smaller value of the metric indicates that the method is more fair.

2) {\bf Language Bias:} We assess language bias by incorporating semantic similarity calculation \citep{simclr,word-fairness} between $prompt(a, c)$ and $prompt('', c)$. Specifically, we use Euclidean distance to evaluate the distance between prompt terms with and without explicit sensitive attributes. The closer distance indicates a potential bias in the language representation towards one specific sensitive attribute. We define our language bias evaluation criteria towards attribute $a_i$ for keyword $c_k$ and our input prompts as $LBias_{a_i}(c_k)$: 
\begin{equation}
\small
    LBias_{a_i}(c_k)=-\|f_j- f\|_2 +\frac{1}{|A|} \sum_{a_j\in A} \|f_j-f\|_2,
\end{equation}
where $\|f_j- f\|_2$ represents the Euclidean distance between the prompt generated with the sensitive term $a_i$ and the keyword $c_k$, compared to the prompt generated with no sensitive term. Thus, the total language bias for keyword $c_k$ and our prompt is $\frac{1}{|A|}\sum_{a_i\in A} LBias_{a_i}(c_k)$.  

Based on the above two metrics, we conduct the following experiment in the occupation keyword set in Appendix \ref{dataset}: 1) First, we calculate language bias on every occupation in the keyword dataset over sensitive attribute gender. 2) Then, for each occupation $c$, we use the following prompt format $prompt('',c)$ for guiding the stable diffusion model to generate 100 images and measure the diffusion bias.  

\begin{figure}[h]
\centering
\includegraphics[width=0.48\textwidth]{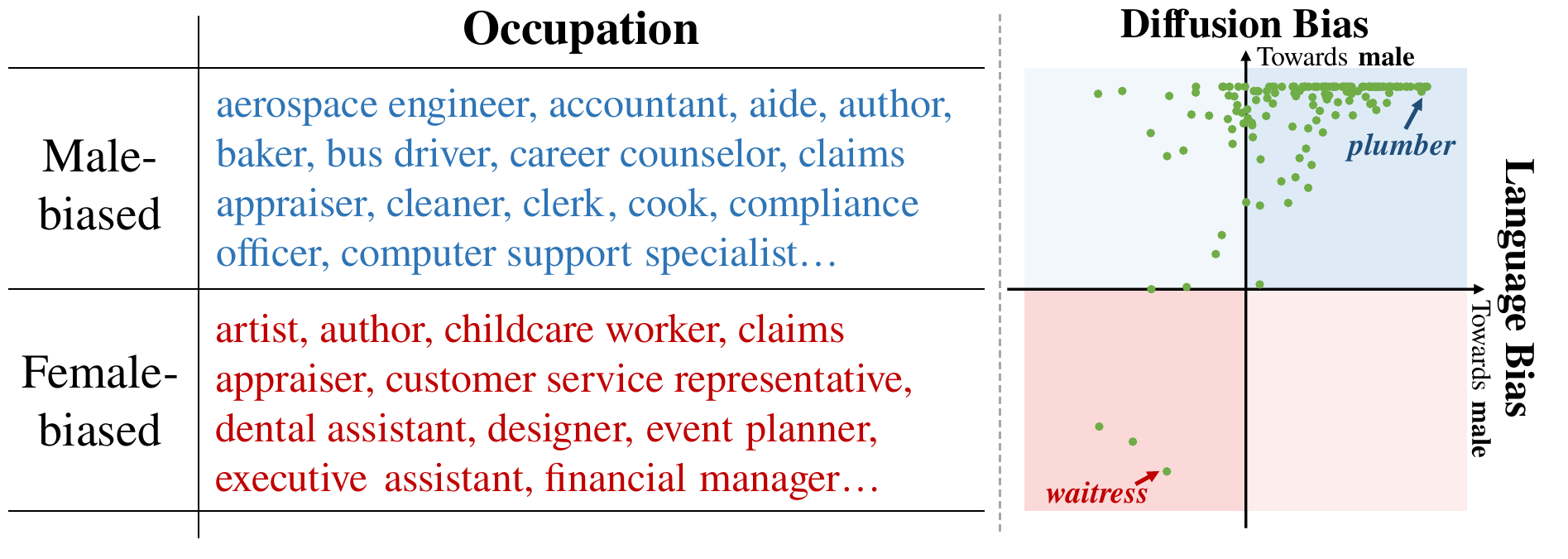}
\vspace{-20pt}
\caption{\textbf{Language Bias and Diffusion Bias Visualization.} We conduct a bias analysis of the language characteristics and the generated outcomes during the diffusion process. Left: Examples of language prejudice. Right: Language bias and diffusion bias for occupational data. Each point represents an occupation. 
} 
\label{fig:bg2}
\vspace{-10pt}
\end{figure}

Figure \ref{fig:bg2} shows the experimental results. In the left region of Figure \ref{fig:bg2}, we conducted a language bias analysis on specific occupations and provided examples to illustrate some samples aligning with societal stereotypes. For example, our analysis of the term "aerospace engineer" highlights a clear gender bias favoring males, reflecting the gender stereotype associated with this profession in the real world. For the right-hand side of Figure \ref{fig:bg2}, We display the biases associated with various occupations by creating a scatter plot with diffusion bias on the y-axis and language bias on the x-axis. Moreover, we discovered the majority of data are concentrated in where a male bias was revealed in both diffusion and language bias. This research implies that language bias and diffusion bias are mutually reinforcing, which infuse in the cross-attention layers of the UNet.
In summary, implicit language bias is one of the direct factors leading to diffusion bias. From the view of language assumption, it is reasonable to reduce the impact it has on diffusion bias and promote more equitable and unbiased generative outcomes.

\section{Mitigating Implicit Bias via Fair Mapping}
\begin{figure}[h]
    \centering
    \includegraphics[width=0.48\textwidth]{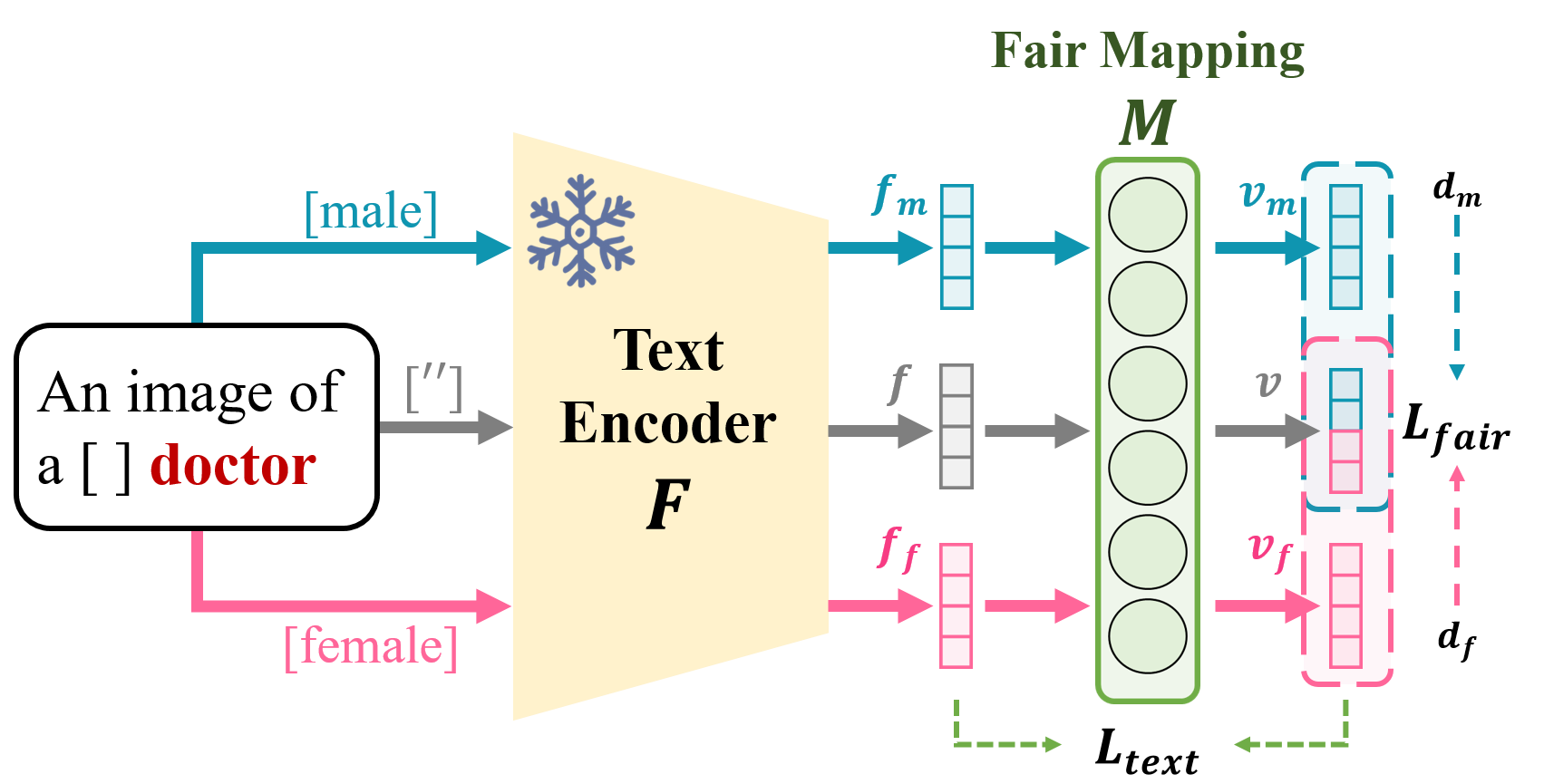}
    \vspace{-10pt}
    \caption{In the training stage, the parameters of the text encoder are frozen, and we apply $\mathcal{L}_{text}$ and $\mathcal{L}_{fair}$ to update Fair Mapping. $d_a$ denotes the distance between $v_a$ and $v$.}
    \label{fig:fw_training}
    \vspace{-10pt}
\end{figure}

In this section, motivated by our above analysis, we introduce our Fair Mapping method to disentangle the implication of language from diffusion generation. 
Generally speaking, Fair Mapping introduces two additional components as a post-processing method on well-trained text-to-image diffusion models: A linear network called Fair Mapping and a detector that will be activated during the inference stage. 
The training and inference procedures of Fair Mapping are elucidated in Figure \ref{fig:fw_training}. We implement Fair Mapping consisting of linear stacking networks, drawing inspiration from StyleGan \citep{stylegan} and MixFairFace \citep{mixfairface}, which enables the correction of native language semantic features, ultimately leading to their debiasing and alignment with the balanced embedding space. 
The detector at the inference stage is used for robust generation to decide whether the input will be debiased. In the following, we will provide details on the training and inference stages. 

\subsection{Training Fair Mapping Network}

Recall that our linear network Fair Mapping, denoted as $M$, is to transform the representation of the input prompt to a debiased space. Our idea is to maintain the representation aligned with a balanced state of sensitive attributes, so they can introduce little assumption to visual representation.
First, given a keyword dataset and group of sensitive attributes, we will construct two distinct types of prompts that have a consistent and uniform format for each keyword $c$. The first type constitutes the original input prompt, denoted as $prompt(\text{` '}, c)$, where we explicitly exclude any sensitive attributes (represented as ` '). It does not prioritize explicit sensitive attribute information. In contrast, the second type of prompt, $prompt(a_j, c)$, where $a_j \in A$, incorporates specific sensitive words. These prompts are designed to quantitatively explore the language relationship between sensitive attributes and the target keyword. Moreover, we have language representation vectors $f$ and $\{f_j\}_{j=1}^{|A|}$ from $prompt(\text{` '}, c)$ and $\{prompt(a_j, c)\}_{a_j\in A}$ respectively. Given these, our Fair Mapping is after the pre-trained text encoder and further transforms these representation vectors:
$$
v=M(f),v_j=M(f_j) \text{ for } a_j\in A.
$$
Our aim with this transformation process is to ensure that the representations are fair and unbiased, and exhibit equitable treatment across sensitive attributes. In detail, the objectives of $v$ and $\{v_j\}_{j}$ are two-fold: 1) They should maintain semantic consistency akin to $f$ and $\{f_j\}_j$ respectively, serving as keeping the original contextual information.
2) More importantly, $v$ should equalize the representation of different demographic groups and prevent the encoding of implicit societal biases.
Therefore, we employ bias-aware objectives and regularization techniques to guide representations to be balanced from sensitive information. Below we will discuss how to achieve the above two goals.

\textbf{Keeping semantic consistency.} Our general idea is designed to maintain consistency and semantic coherence between the original embeddings and mapped embeddings. To achieve this objective, we employ a strategy for minimizing the disparity between pre-transformed and post-transformed features in the embedding space. Specifically, we adopt the mean squared error (MSE) as a metric to measure the reconstruction error, drawing inspiration from the reconstruction rule proposed in decoder architecture \citep{vae}. By applying this metric, we compute a semantic consistency loss for each keyword:
\begin{equation}
\label{eq:consis}
\small
    \mathcal{L}_{text}= \frac{1}{|A|+1} \left( ||v-f||_2^2+ \sum_{a_j \in A} ||v_j-f_j||_2^2 \right). 
\end{equation}
Through the minimization of this loss over all keywords, we can ensure that the mapped embeddings preserve the crucial information and semantic attributes inherent in the original embeddings. This process safeguards the fidelity and integrity of the data throughout the mapping transformation. 

\textbf{Fairness loss.} Distances of embeddings to groups with sensitive attributes can inadvertently encode demographic information. For example, if the word ``doctor'' is closer to ``male'' than ``female'' in the representation space of language models, it may inherently convey gender bias \citep{simclr}. To mitigate this issue, we employ an invariant loss that entails the adjustment of associations between sensitive attributes and naive prompts during the training process using mapping offsets. The primary objective is to diminish associations of implicit sensitive attributes with text embeddings through the application of mapping offsets, promoting a more neutral and unbiased representation.

To equalize the representations of attributes, we ensure that representations of prompts have balanced distances from the representations of prompts with specific sensitive attributes, thereby reducing the bias assumptions in the semantic space. 
In the case where the size of the sensitive group $A$ is $2$, we can minimize the difference in distance between the native embeddings, expressed as $|d(v,v_1)-d(v,v_2) |$. Here, $d(\cdot,\cdot)$ represents the Euclidean distance \citep{distance-euc} between embeddings. To address the computational complexity when dealing with a large attribute set containing multiple sensitive groups, we can optimize representation bias by reducing the variance in the distance between the embeddings instead of calculating the difference in the distance for each pair. The fairness loss term, denoted as $\mathcal{L}_{fair}$, can be formulated as follows:
\begin{equation}
\small
\label{eq:fair}
\mathcal{L}_{fair} = \sqrt{\frac{1}{|A|} \sum_{a_j \in A} \left( d(v,v_j) - \overline{d(v,\cdot)} \right)^2}.
\end{equation}
Here, $d(v,v_i)$ represents the Euclidean distance between the native embedding $v$ and the specific sensitive attribute embedding $v_i$. $\overline{d(v,\cdot)}$ refers to the average distance between the native embedding $v$ and all the sensitive attribute embeddings $v_j$. By incorporating this fairness loss term into the training objective, we aim to minimize the variance in the distance between $v$ and $v_j$ with sensitive attributes. To optimize the overall objective, we combine the semantic consistency loss, denoted as $\mathcal{L}_{text}$ (from (\ref{eq:consis})), with the estimated bias difference from the fairness penalty (from  (\ref{eq:fair})). This results in the following combined loss function for each keyword:
\begin{equation}
\label{eq:total}
\mathcal{L} = \mathcal{L}_{text} + \lambda \mathcal{L}_{fair}, 
\end{equation}
where $\lambda$ is a hyperparameter that controls the trade-off between semantic consistency and fairness. By minimizing this combined loss function, we aim to simultaneously ensure semantic consistency and reduce bias in the language representation. We optimize the parameters of Fair Mapping while keeping the parameters of the diffusion model fixed.
\vspace{-0.1in}
\subsection{Inference}

\begin{figure}[h]
    \centering
    \includegraphics[width=0.45\textwidth]{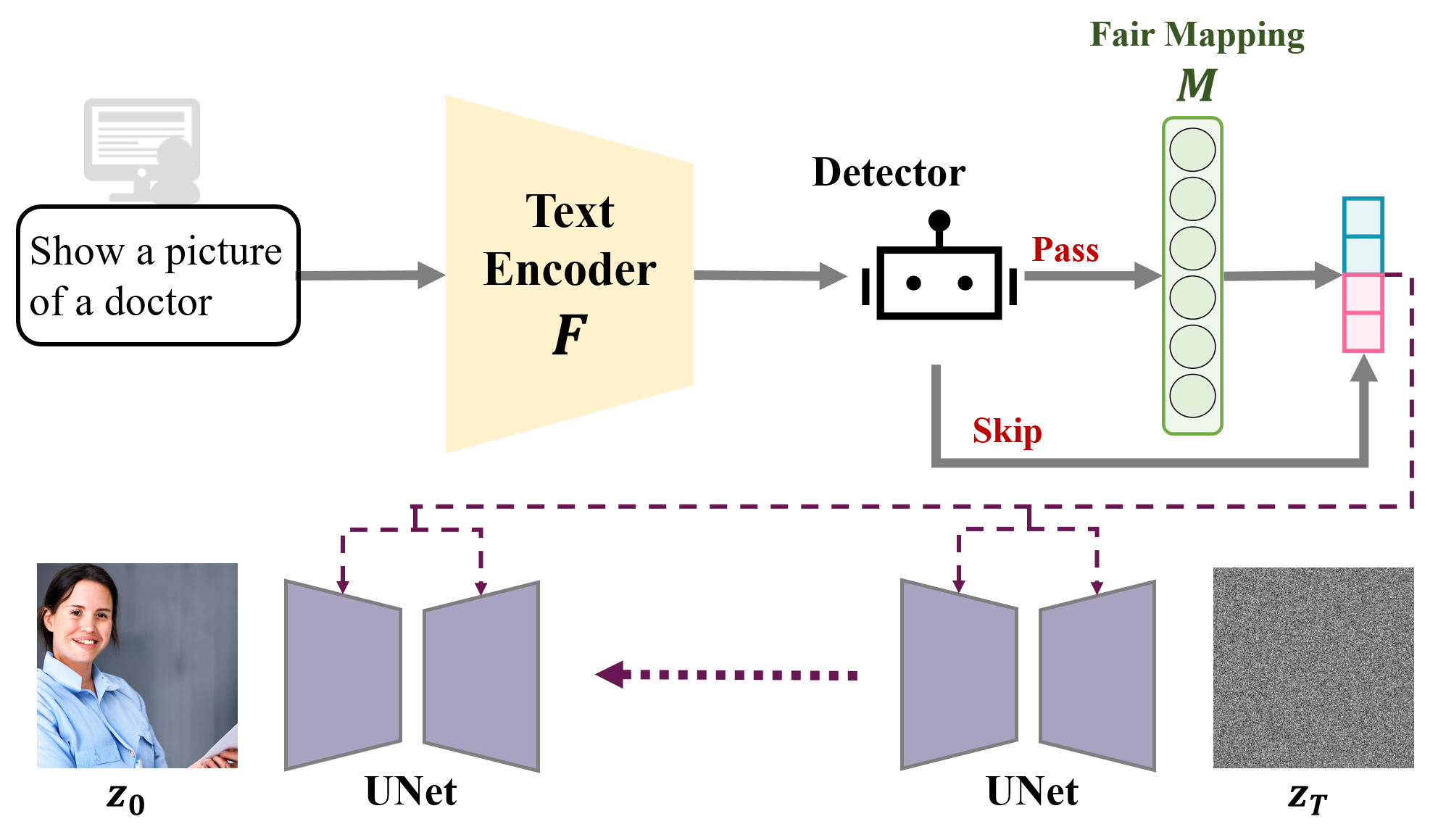}
    \vspace{-10pt}
    \caption{In the inference stage, the detector after the text encoder determines whether the text should pass or skip the Fair Mapping linear network.}
    \label{fig:fw_inference}
\vspace{-0.1in}
\end{figure}
In the inference stage, as Figure \ref{fig:fw_inference} demonstrated, the detector is introduced before Fair Mapping. To build a robust system for users, the main goal of the detector is to decide whether the text should pass or skip Fair Mapping. In detail, it aims to match the target input with implicit sensitive attributes and avoid solving with explicit sensitive attributes. Due to the space limit, the details for the detector can be found in Appendix \ref{sec:inference} and Algorithm \ref{alg:text-processing}.

\indent {\bf Discussions.}  We can easily see that there are several strengths of Fair Mapping. 
Firstly, Fair Mapping is model-agnostic, i.e., as it only introduces a linear mapping network and a detector after the text encoder, it can easily be integrated into any text-to-image diffusion model as well as be a plug-in text-encoder with the same parameters.
Secondly, Fair Mapping is lightweight. As a post-processing approach, Fair Mapping only introduces an additional linear map to be optimized while keeping the parameters of the diffusion model fixed. Moreover, as we will mention in the experiments, an eight-layer linear network is sufficient to achieve good performance on both utility and fairness with little additional time cost. 
Finally, our method is quite flexible. Due to the simplicity of our loss for each keyword, our linear network can be customized for any other prompts, loss of semantic consistency, and loss of fairness.

\section{Experiments}

\begin{figure*}[h]
\centering
\includegraphics[width=0.9\textwidth]{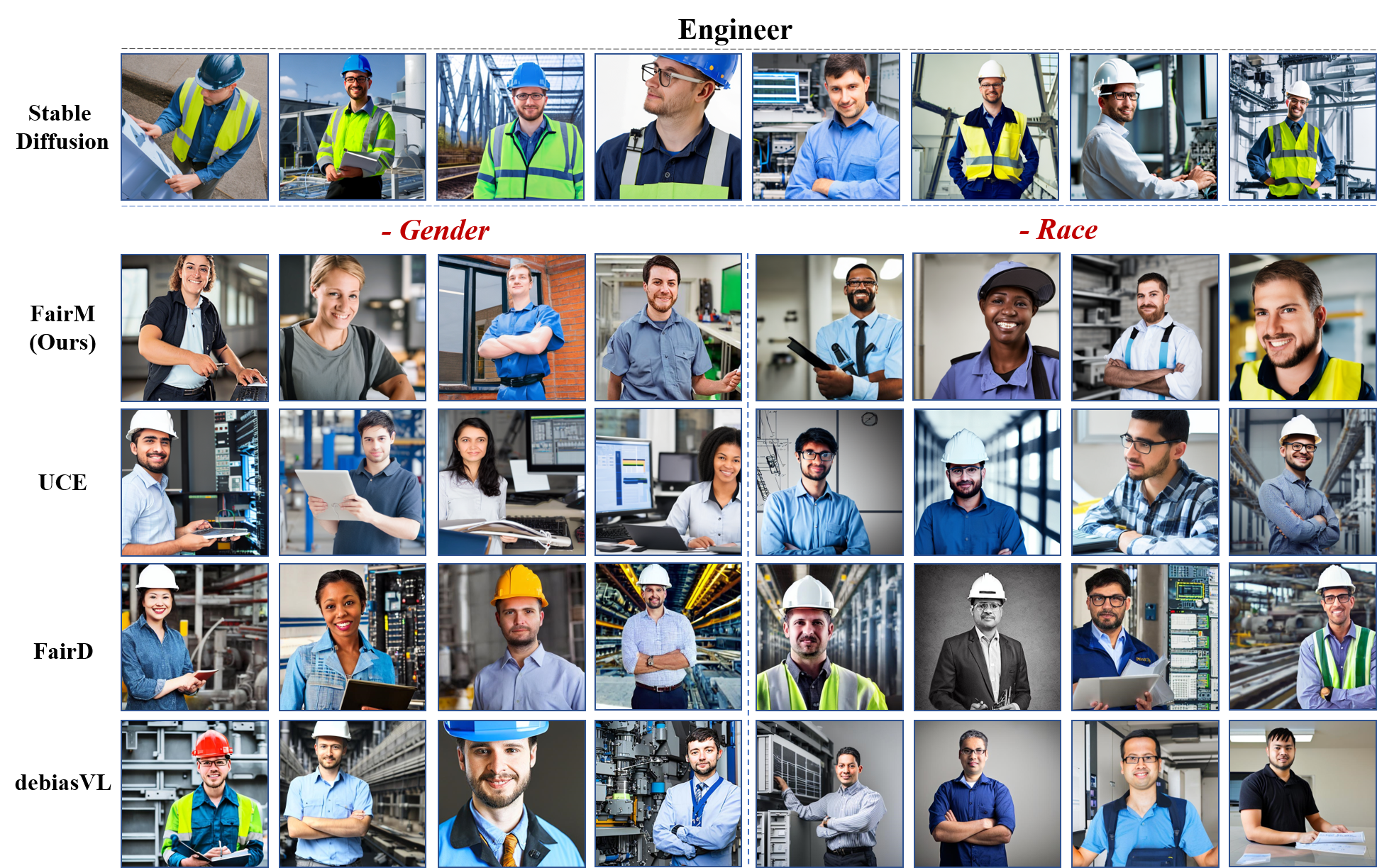}
\caption{Comparison with original SD and different debiasing methods in prompt "an image of an engineer". Our method makes generated images equally represent genders and races. More visual results are in Appendix \ref{app-visual}.}
\label{fig:exp-visual}
\end{figure*}

We mainly compared our Fair Mapping(FairM) with three text-guided diffusion models: Stable Diffusion (SD) \citep{ldm}, Structured Diffusion (StruD) \citep{structure-diffusion} and Composable Diffusion (ComD) \citep{codi}, as well as three state-of-the-art methods for fair diffusion generation: Fair Diffusion (FairD) \citep{fairdiffusion}, Unified Concept Editing (UCE) 
 \citep{fair_editing_unified} and Debias Visual Language (debiasVL) \citep{fair_language_chuang2023debiasing}.
We report the performance of our models from three aspects: Our Fair Mapping 1) outperforms baselines in fairness evaluation and computation cost, 2) showcases alignment and diversity in human-related descriptions effectively by quantitative analysis, 3) is more lightweight and introduces acceptable time overhead 4) matches human preferences in image quality and text alignment of the state-of-the-art text-to-image diffusion methods.

\subsection{Experimental Setup}

\textbf{Datasets.} We select a total of 150 occupations and 20 emotions for the fair human face image generation following \citep{diffusion-bias}. For sensitive groups, we choose gender groups (male and female) and racial groups (Black, Asian, White, Indian) provided by \citep{dataset-fairface}. We provide a comprehensive list of keywords in Appendix \ref{dataset}.

\textbf{Implementation details.} In our experiments, we use Stable Diffusion \citep{ldm} trained on the LAION-5B \citep{dataset-laion5b} dataset as the base model and implement 50 DDIM denoising steps for generation. Specifically, we utilize the pre-trained stable diffusion model (SD-1.5). All of our training experiments are conducted on one Nvidia A100 GPU. We maintain a uniform learning rate of 1e-2 and keep the number of training epochs consistent at 500. For each specific occupation and emotion, we set $\lambda$ to 0.1. We set the number of layers to eight for linear mapping structure in Fair Mapping. For the prompts for training and inference, for the occupation keyword set, we employ a standardized format in the form of $prompt(a, c)$=``an image of an $a$ $c$'' as the prompt. Similarly, for the emotion keyword set, we utilize a consistent format of $prompt(a, c)$=``an image of an $a$ $c$ person'' as the prompt.

\textbf{Evaluation metrics.} 
\label{metric}
Besides the two evaluation metrics in the above Section \ref{sec:language_bias}, language bias and diffusion bias (we denote as Bias), below we introduce some metrics for the utility and the quality of the generated images.

1) {\bf Alignment:} To measure the alignment between generated images and human-related content, we adopt the CLIP-Score \cite{metri_clip_score}, which measures the distance between input textual features and generated image features. Due to the limitation \citep{metric-bias} of capturing the specific requirements of human-related textual generation, we introduce the Human-CLIP metric (see Appendix \ref{human-clip} for details), which focuses on evaluating the CLIP-score related to human appearance. 

2) {\bf Diversity:} We use intra-class average distance (ICAD) \citep{intra_class} to evaluate the diversity of generative results. For each keyword, we measure the average distance between all generated images and the center of images by using a distance metric by squared Euclidean distance (see Appendix \ref{diversity} for details). A lower intra-class average distance suggests that the generative results are more similar and excessively focus on a few specific samples.

\subsection{Experimental Results}

Figure \ref{fig:exp-visual} demonstrates the effectiveness of our method for debiasing. When compared to existing text-guided diffusion approaches, Fair Mapping demonstrates an enhanced capacity for generating diverse sensitive groups while upholding the stability of the generated results. 

\begin{figure}[h]
\centering
\includegraphics[width=0.48\textwidth]{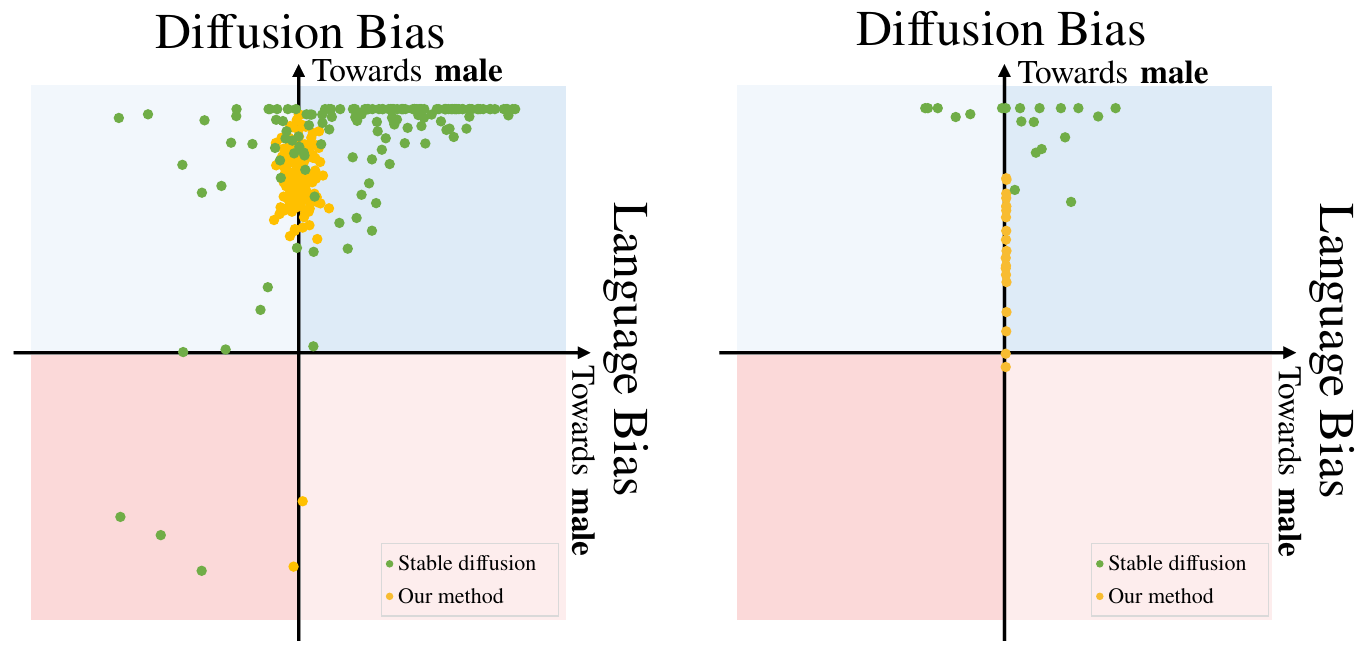}
\vspace{-20pt}
\caption{\textbf{The comparative evaluation between our method and Stable Diffusion on language bias and diffusion bias}. Each data point represents a distinct description. Our method reduces bias in generated language representations and achieves a more equitable distribution of generated images.}
\label{fig:fair_point}
\vspace{-5pt}
\end{figure}

\textbf{Fair Mapping can mitigate bias in diffusion models.} The comparative analysis depicted in Figure \ref{fig:fair_point} underscores the significant strides made by our method in mitigating language bias and diffusion bias. In stable diffusion, data points display a broad distribution of language biases, favoring the male gender in the generated images as indicated by higher diffusion bias values (y-axis). In contrast, our approach achieves a more concentrated set of results with language biases around zero and a distribution of relatively low diffusion biases, which shows our methods effectively mitigate implicit language bias in text embeddings and promote balanced visual representation in generated images. 

Besides a general illustration of language bias and diffusion bias. 
Table \ref{tb:fairness} showcases the fair evaluation results for sensitive attributes: gender and race. Fair Mapping demonstrates significant improvements in fairness compared to other baseline approaches. It achieves lower Diffusion Bias compared to Stable Diffusion for both gender (18\% for Occupation and 54\% for Emotion) and race (14\% for Occupation and 38\% for Emotion) datasets. Besides, compared with other debiasing methods, Our method outperforms in enhancing the representation of minority groups. We did not test for Fair Diffusion primarily because the generated results heavily depend on subjective post-processing by humans.

\vspace{-12pt}
\begin{table}[h]
\renewcommand{\arraystretch}{0.7}
\centering
\caption{Fair evaluation results of sensitive group gender and race. O denotes the Occupation dataset and E denotes the Emotion dataset. The \textbf{bold} value denotes the best performance.}
\resizebox{0.8\linewidth}{!}{
\begin{tabular}{cccc}
\toprule
Attribute  &  Models & Bias (O) & Bias (E) \\
\midrule
\multirow{6}{*}{Gender} & SD     & 0.4466 & 0.4652 \\
                        & StruD  & 0.4141 & 0.4100    \\
                        & ComD & 0.4027 & 0.4203 \\
                        &  UCE &0.3802 & 0.3266 \\
                        & debiasVL &0.4573 & 0.4178\\
                        & FairM(Ours) & \textbf{0.3625} & \textbf{0.2113} \\
\midrule
\multirow{6}{*}{Race}   & SD     & 0.2599 & 0.1893 \\
                        & StruD  & 0.2368 & 0.1824 \\
                        & ComD & 0.2344 & 0.1489 \\
                        &  UCE & 0.2314 & 0.1505\\
                        & debiasVL & 0.2992 & 0.1592\\
                        & FairM(Ours) & \textbf{0.2231} & \textbf{0.1178} \\
\bottomrule
\end{tabular}
}
\label{tb:fairness}
\vspace{-7pt}
\end{table}

\textbf{Alignment and diversity.} In Table \ref{tab:exp2}, when comparing with baselines for text-to-image diffusion generation, debiasing methods restrict the model's ability to generate images as a trade-off for fair generation. Although both UCE methods maintain the quality of generation, they perform poorly in Human-CLIP evaluation.
Furthermore, Our method demonstrates a strong alignment with human-related prompts, with 13\% improvements in the Human-CLIP metric for occupation. our method successfully captures human-related characteristics in the generated images, despite having a slight loss in CLIP-Score. 

\begin{table}[h]
\vspace{-10pt}
\renewcommand{\arraystretch}{1.2}
\centering
\caption{Evaluation results of image alignment and diversity. CLIP denotes as CLIP-Score and CLIP-H denots as CLIP-Human. The \textbf{bold} value denotes the best performance.\label{tab:exp2}}
\resizebox{0.95\linewidth}{!}{
\begin{tabular}{ccccccc}
\toprule
\multirow{2}{*}{Models} & \multicolumn{3}{c}{Occupation} & \multicolumn{3}{c}{Emotion} \\
\cmidrule{2-4} \cmidrule{5-7} 
&CLIP & CLIP-H  & ICAD & CLIP & CLIP-H & ICAD\\
\midrule
SD     &0.2320 & 0.1339 &  \textbf{3.64}       & \textbf{0.2026} & \textbf{0.1399}       & \textbf{3.84} \\
StruD  & \textbf{0.2339} & 0.1284  & 3.61 & 0.1930 &0.1103             & 3.82 \\
ComD &0.2299 &\textbf{0.1367}          &  3.60        & 0.1901   & 0.1155          & 3.82  \\
\hline
FairD-Gender   & \textbf{0.2274} &  0.1348 & 3.62 & \textbf{0.1894} &  0.1298 & \textbf{3.91} \\  
UCE-Gender & 0.2280 & 0.1133 & 3.61  &0.1848  & 1.1198          & 3.76 \\
dibiasVL-Gender & 0.2011 & 0.1223 &  3.47 & 0.1806 & 0.0890         & 3.79  \\
\rowcolor{grey!20}
FairM-Gender   & 0.2021 &  \textbf{0.1494}       & \textbf{3.69} & 0.1809&  \textbf{0.1366}          & \textbf{3.91}  \\ 
\hline
FairD-Race    & \textbf{0.2239} &   0.1292    &  3.64  & 0.1882 & 0.1266           & 3.89   \\
UCE-Race & 0.2327 & 0.1045&  3.58 & \textbf{0.1914}  &0.0968           & 3.77 \\
dibiasVL-Race & 0.2187&0.1022 & \textbf{3.68}  & 0.1795 &  0.1083         & 3.77 \\
\rowcolor{grey!20}
FairM-Race          & 0.2197 &   \textbf{0.1522}      &  \textbf{3.68}        &0.1848 & \textbf{0.1324}           & \textbf{3.95}   \\ 
\bottomrule
\end{tabular}
}
\end{table}

Regarding diversity, for the Occupation dataset in Table \ref{tab:exp2}, our method demonstrates a significant improvement in ICAD of generated results when compared to Stable Diffusion. Specifically, the metrics show an increase from 3.64 to 3.68 and 3.69 for gender and race groups, respectively. For the Emotion dataset, the ICAD of our method is still better than other debiasing text-to-image methods. These results demonstrate our method's ability to excel in generating varied results in diverse environments. 

\textbf{Computation cost.} On an Nvidia V100 device, our method finishes training for debiasing on 150 occupations in just 50 minutes, which demonstrates impressive efficiency for training. Table \ref{tb:fd_time} offers comparisons of the time for generating 100 images. Our method showcases a commendable performance, generating 100 images for a single occupation in 434 seconds, which is close to SD. UCE increases the minimal time in inference due to adjusting the origin parameters, but it requires a significant amount of training time.

\begin{table}[h]
\vspace{-10pt}
\renewcommand{\arraystretch}{1}
\centering
\caption{Evaluation Results in time consumption on generation of 100 images.\label{tb:fd_time}}
\resizebox{0.5\linewidth}{!}{
\begin{tabular}{ccc}
\toprule
Models & time(seconds)\\
\midrule
SD   & 424    \\
\hline
FairD & 1463 \\
UCE & 426 \\
debiasVL & 764 \\
\rowcolor{grey!20}
FairM(ours) & 434\\
\bottomrule
\end{tabular}
}
\end{table}

\textbf{Human preference.} In Table ~\ref{tab:human}, we also conduct a human study about the fidelity and alignment of our method. For fidelity, the human preference scores reveal that our method consistently outperforms the other generative images both for occupation and emotion description. 
As Fair Mapping introduces a trade-off between prioritizing fairness and maintaining the alignment of facial expressions and textual descriptions, some participants expressed dissatisfaction with our method's performance in achieving consistency between facial expressions and textual descriptions. 
Future research may focus on enhancing the consistency of the text prompts while balancing the bias. 

\begin{table}[h]
\vspace{-10pt}
\renewcommand{\arraystretch}{0.8}
\centering
\caption{Evaluation Results in Human Preference. The higher the score, the more it aligns with human preferences. Please refer to the Appendix \ref{human} for details.\label{tab:human}}
\resizebox{\linewidth}{!}{
\begin{tabular}{ccccc}
\toprule
\multirow{2}{*}{Models} & \multicolumn{2}{c}{Occupation} & \multicolumn{2}{c}{Emotion} \\
\cmidrule{2-3} \cmidrule{4-5} 
&Fidelity  & Alignment & Fidelity & Alignment\\
\midrule
SD     & 2.7558          &  \textbf{3.6760}        & 2.7230       & 3.4929 \\
StruD  & 2.5399          &  2.9953          & 3.3427      & 3.3615 \\
ComD & 2.6667          &  3.0375          & 1.9718          & \textbf{3.6854} \\
\midrule
FairM-Gender          &  3.0140         & 3.0760         &  3.4883          & 3.2431 \\ 
FairM-Race          &    3.0798       &  3.3661        & 3.0140           & 3.3475 \\ 
\midrule
Real Image &   \textbf{3.4694}        &     -      & \textbf{3.5576}          & - \\ 
\bottomrule
\end{tabular}
}
\vspace{-7pt}
\end{table}

\subsection{Ablation Study} 


Finally, we conduct an ablation study on the necessities of our two components $\mathcal{L}_{text}$ in Eq.\ref{eq:consis} and $\mathcal{L}_{fair}$ in loss (\ref{eq:fair}). Table \ref{tab:ab} shows the results of an ablation study examining the influence of different factors in the loss terms on model performance. We implement our experiments on sensitive attribute groups {Gender} and dataset {Occupation}. First, we can see that $\mathcal{L}_{text}$ can function independently, as indicated by individual rows representing the method's performance when only one of these criteria is considered. However, it is evident that $\mathcal{L}_{fair}$ alone is not effective, indicating the necessity of $\mathcal{L}_{text}$ to establish an effective semantic space. Secondly, we can observe the combination of generating diverse sensitive attributes ($\mathcal{L}_{text}$) and maintaining fairness in representation ($\mathcal{L}_{fair}$) achieves the lowest diffusion bias, indicating the superior performance of fairness.


\begin{table}[h]
\vspace{-15pt}
\renewcommand{\arraystretch}{0.7}
\centering
\caption{An ablation study on $\mathcal{L}_{fair}$ and $\mathcal{L}_{text}$ in the loss function. O denotes the Occupation dataset and E denotes the Emotion dataset.}
\resizebox{0.7\linewidth}{!}{
\begin{tabular}{cccc}
\toprule
$\mathcal{L}_{text}$ & $\mathcal{L}_{fair}$ & Bias(O) &Bias(E)  \\
\midrule
- & - & 0.4466 & 0.4622 \\
- & $\checkmark$ & - & -  \\
$\checkmark$ & - & 0.4030 &0.3862 \\
$\checkmark$ & $\checkmark$ & \textbf{0.3624} & \textbf{0.2113}  \\
\bottomrule
\label{tab:ab}
\end{tabular}%
}
\vspace{-7pt}
\end{table}

\begin{figure}[h]
    \centering    \includegraphics[width=0.4\textwidth]{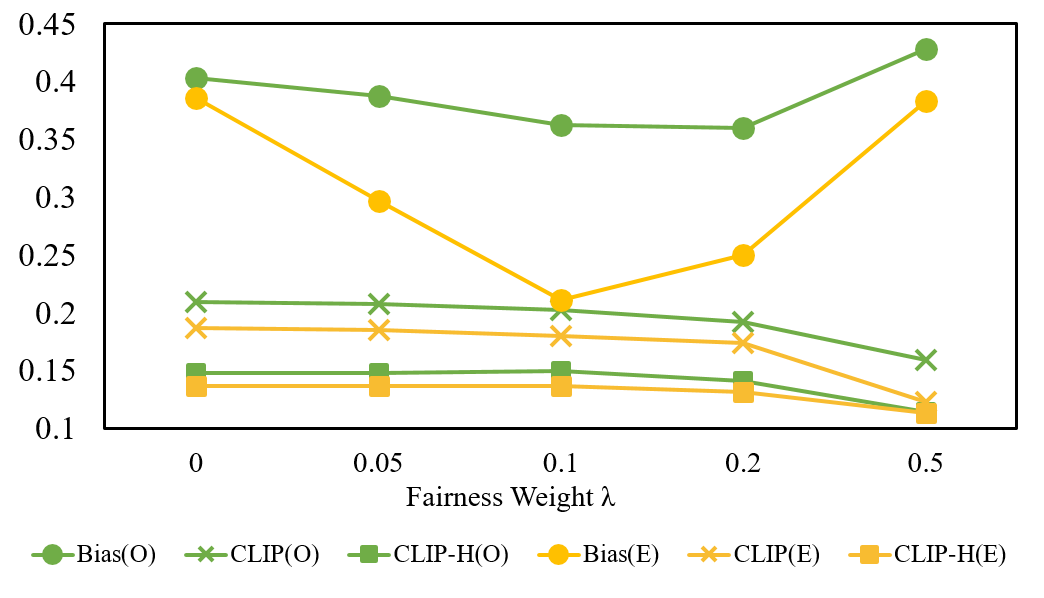}
    \vspace{-10pt}
    \caption{\textbf{The influence of different regularization parameter $\lambda$}. (O) means Occupation dataset and (E) means Emotion dataset.}
    \label{fig:lambda}
\end{figure}

In Figure \ref{fig:lambda} we study the effect of the fairness penalty regularization parameter $\lambda$. Firstly, we can see that when $\lambda$ becomes larger, both CLIP-Score and Huma-CLIP decrease. This is due to a larger $\lambda$ implying that we will more focus on fairness rather than the quality of the generated images.  Moreover, these two metrics only slightly decrease when $\lambda$ is less than $0.1$, which further supports our previous conclusion that our method has almost the same quality of generated images. Secondly,  we can observe that when $\lambda$ is larger and smaller than $0.1$, the BiasScore will decrease. However, when $\lambda$ is greater than $0.1$, the BiasScore will increase. We think in this case, the generated images experience severe distortion, resulting in a reduced amount of semantic information and consequently leading to a decline in fairness as well.

\vspace{-7pt}
\section{Conclusions}
In this paper, we advocated that the implicit biases in the input text prompt contribute to significant observed bias in current text-to-image diffusion models. We developed Fair Mapping, a model-agnostic debiasing mapping network, to effectively mitigate bias with few additional parameters for training. Meanwhile, it is flexible for Fair Mapping to adapt to different customized data.
Furthermore, in our fairness evaluation metrics, experiments demonstrate substantial efficiency compared to text-guided diffusion models and other debiasing methods.

\section{Broader Impact}
Generative models, including large language models and image generation models, learn statistical regularities from input data to generate new content. There are potential risks if the training data contains biases, such as the generated content is likely to reflect those stereotypes. By addressing potential biases in text-to-image model generation, our work contributes to ensuring that the generated images are free from biases related to gender, race, or other social attributes. It paves the way for future research and applications to ensure that AI-generated content breaks stereotypes and discrimination in society, promoting more equitable and inclusive opportunities.

\bibliography{reference}
\bibliographystyle{icml2024}

\newpage
\appendix
\onecolumn

\section{Details of the Inference Stage}\label{sec:inference}

In the inference stage, Fair Mapping should keep robustness to meet requirements of possible debiasing content. For example, Fair Mapping should be activated whatever the user's prompt could be ``I want to show a $c$ figure'', ``An image of a $c$'' and other formats containing keywords with implicit bias. To ensure reliability across diverse descriptions, an additional detector is introduced with the primary objective of adapting the user's input prompt to a training prompt that exhibits the closest semantic similarity. To achieve this, we calculate the similarity distance between the input prompt and each training prompt of the linear network using a pre-trained text encoder. Subsequently, we identify the training prompt that exhibits the smallest distance. If the calculated distance falls below a pre-defined threshold, we transform the input prompt to match the identified training prompt.

However, an additional issue may arise. The linear network aims to debias the implicit bias associated with the prompt without explicit biased information. It can easily misunderstand explicit biased information in the input text, such as "An image of a male doctor", where the use of the linear network becomes unnecessary. Therefore, it becomes imperative for the detector to identify the presence of any sensitive attribute in the transformed training prompt. If the closest training prompt lacks sensitive attributes, passing it through the linear network for debiasing becomes necessary. Conversely, if sensitive attributes are present, skipping the linear network is warranted. The detailed algorithm for the detector is provided in Algorithm \ref{alg:text-processing}.

\begin{algorithm}[H]
	\SetAlgoLined
	\KwIn{Input textual prompt $w$, a threshold $e>0$, a keyword training set $C$ with sensitive attribute set $A$, training prompt set $S$ for $C$ and $A$}
	\KwOut{Modified text}
         \For{Each prompt $s\in S$}{ 
	Calculate similarity distance $d_s$ = SimilarityDistance($w$, $s$)\;
 } 
        $id=\arg\min_{s\in S}(d_s)$\;
        $tp$=$S$[$id$]\;
        
        \If{$d<e$}{
            \If{$tp$ does not contain any sensitive attribute word in $A$}{
    Return $w$ to the text encoder and skip the linear network\; 
			}
            \Else{
            Return $tp$  to the text encoder\;
            }
        }
	Return $w$ to the text encoder and skip the linear network\;
	
	\caption{Detector Processing Algorithm}
	\label{alg:text-processing}
\end{algorithm}

\section{Preliminary}
Diffusion models generate images by iteratively a series of denoising steps, progressively reducing the noise and improving the image quality. The process involves gradually refining the initial noise map to produce high-quality images. A predefined number of denoising steps determines the degree of noise at each step and a timestep-dependent noise prediction network $\epsilon_{\theta}$ is trained to predict the noise added to a given initial noise map as input $z$. Although earlier models, such as Denoising Diffusion Probabilistic Models (DDPM) \citep{ddpm}, are computationally expensive, the non-Markovian diffusion method Denoising Diffusion Implicit Models (DDIM) \citep{ddim}, has improved the inference speed by drastically reducing the number of denoising steps. In DDIM, the noise prediction network $\epsilon_{\theta}$ is utilized to estimate the noise added at each denoising step. By reducing the number of denoising steps, DDIM achieves faster inference without compromising the quality of generated images. This improvement in computational efficiency allows for more practical and scalable implementation of diffusion models in various applications.

Text-to-image diffusion models involve the utilization of diffusion models in combination with textual descriptions to generate image samples. The goal of this process is to produce images that correspond to the given conditional textual information, represented as $pt$, thereby establishing an explicit and controllable condition for image generation. According to the sampling process of DDIM with condition $pt$,
\begin{equation}
\small
\boldsymbol{x}_{t-1}=\sqrt{\alpha_{t-1}} \underbrace{\left(\frac{\boldsymbol{x}_t-\sqrt{1-\alpha_t} \epsilon_\theta^{(t)}\left(\boldsymbol{x}_t \mid pt\right)}{\sqrt{\alpha_t}}\right)}_{\text {predicted } \boldsymbol{x}_0 }+\underbrace{\sqrt{1-\alpha_{t-1}} \cdot \epsilon_\theta^{(t)}\left(\boldsymbol{x}_t \mid pt\right)}_{\text {direction pointing to } \boldsymbol{x}_t },
\end{equation}

Where $\boldsymbol{x}_{t-1}$ represents the previous sample at time step $t-1$, $\alpha_{t-1}$ denotes the diffusion hyperparameter at time step $t-1$, $\boldsymbol{x}_t$ is the current sample at time step $t$, $\epsilon_\theta^{(t)}\left(\boldsymbol{x}_t \mid pt \right)$ denotes the noise added to the current sample at time step $t$, which is parameterized by $\theta$ and conditioned on $pt$. $\epsilon_\theta^{(t)}\left(\boldsymbol{x}_t \mid pt\right)$ holds the key to generate images with conditions.

By applying Bayes' theorem, we can deconstruct the conditional generation probability as it relates to the generation of $\boldsymbol{x}_t$ given $\mathbf{pt}$:
$p(\boldsymbol{x}_t \mid \mathbf{pt}) = \frac{p(\mathbf{pt} \mid \boldsymbol{x}_t) p(\boldsymbol{x}_t)}{p(pt)}.$
As a result, the gradient of the logarithm of the conditional generation probability $\log p\left(\boldsymbol{x}_t \mid pt \right)$ can be decomposed as follows:

\begin{equation}
\begin{aligned}
\small
\nabla \log p\left(\boldsymbol{x}_t \mid pt\right) 
& =\nabla \log p\left(\boldsymbol{x}_t\right)+\nabla \log p\left(pt\mid \boldsymbol{x}_t\right)-\nabla \log p(pt) \\
& =\underbrace{\nabla \log p\left(\boldsymbol{x}_t\right)}_{\text {unconditional score }}+\underbrace{\nabla \log p\left(pt \mid \boldsymbol{x}_t\right)}_{\text {classifier gradient }}
\label{eq:classifier-guidance}
\end{aligned}
\end{equation}

The decomposition of the log gradient of the conditional generation probability $\log p\left(\boldsymbol{x}_t \mid pt\right)$ yields the unconditional score $\log p\left(\boldsymbol{x}_t\right)$ and the classifier gradient$\log p\left(pt \mid \boldsymbol{x}_t\right)$, as well as a marginal probability $\log p(pt)$. We can see that the log gradient of the conditional generation probability in the conditional diffusion model can be decomposed into the sum of the unconditional score and the classifier gradient. This decomposition allows for considering the influence of both the unconditional probability and the classifier gradient during the generation process. It can be achieved by Classifier Guidance by using a classifier model to provide additional guidance or constraints during the generation process. However, the utilization of an explicit classifier for classifier guidance in conditional generation presents several challenges. Firstly, it necessitates the training of an additional classifier capable of accommodating noisy inputs. Secondly, the efficacy of generating samples aligned with specific categories can be influenced by the quality of the classifier. Lastly, the generated images may deceive the classifier by exploiting minute details, leading to samples that deviate from the intended conditional specifications.

On the contrary, the Classifier-Free Guidance method balances the realism and diversity of generated images by modifying the guidance weights. The method replaces the explicit classifier with an implicit classifier, which removes the necessity for direct computation of both the explicit classifier and its gradients:
\begin{equation}
\begin{aligned}
\small
\nabla \log p\left(pt \mid \mathbf{x}_t\right) & =\nabla \log p\left(\mathbf{x}_t \mid pt\right)-\nabla \log p\left(\mathbf{x}_t\right) \\
& =-\frac{1}{\sqrt{1-\bar{\alpha}_t}}\left(\boldsymbol{\epsilon}_\theta^{(t)}\left(\mathbf{x}_t, pt\right)-\boldsymbol{\epsilon}_\theta^{(t)}\left(\mathbf{x}_t\right)\right)
\label{eq:gradient}
\end{aligned}
\end{equation}

The gradient expression in (\ref{eq:gradient}) demonstrates how the guidance is achieved without utilizing an explicit classifier. Instead, the gradient is obtained by subtracting the noise term $\boldsymbol{\epsilon}_\theta^{(t)}\left(\mathbf{x}_t\right)$ from the noise term conditioned on the target condition $\boldsymbol{\epsilon}_\theta^{(t)}\left(\mathbf{x}_t, pt\right)$. This gradient helps balance the trade-off between realism and diversity in the generated images. The new generation process no longer relies on an explicit classifier. Substituting the classifier gradients from Equation \ref{eq:gradient} and our origin $\boldsymbol{\epsilon}_\theta^{(t)}\left(\mathbf{x}_t, pt\right)$ can be replaced by $\overline{\boldsymbol{\epsilon}}_\theta^{(t)}\left(\mathbf{x}_t, pt\right) $:

\begin{equation}
\begin{aligned}
\overline{\boldsymbol{\epsilon}}_\theta^{(t)}\left(\mathbf{x}_t, pt\right) & =\boldsymbol{\epsilon}_\theta^{(t)}\left(\mathbf{x}_t, pt\right)-\sqrt{1-\bar{\alpha}_t} w \nabla_{\mathbf{x}_t} \log p\left(pt \mid \mathbf{x}_t\right) \\
& =\boldsymbol{\epsilon}_\theta^{(t)}\left(\mathbf{x}_t, pt\right)+w\left(\boldsymbol{\epsilon}_\theta^{(t)}\left(\mathbf{x}_t, pt\right)-\boldsymbol{\epsilon}_\theta^{(t)}\left(\mathbf{x}_t\right)\right) \\
& =(w+1) \boldsymbol{\epsilon}_\theta^{(t)}\left(\mathbf{x}_t, pt\right)-w \boldsymbol{\epsilon}_\theta^{(t)}\left(\mathbf{x}_t\right)
\end{aligned}
\end{equation}

$w$ is a scale for conditioning. As a result, the Classifier-Free Guidance method fits a conditional $\boldsymbol{\epsilon}_\theta^{(t)}\left(\mathbf{x}_t, pt\right)$ and an unconditional $\boldsymbol{\epsilon}_\theta^{(t)}\left(\mathbf{x}_t\right)$ simultaneously. The unconditional generation process generates text samples without any specific constraints or guidance, and it is responsible for the diversity of the generated outputs. On the other hand, the conditional generation process generates image samples that are guided by specific inputs of the condition. In the process of diffusion-based conditional generation, the resulting images are influenced by both the conditioning information and the latent prior. By combining the conditioning information with the latent prior, the generative model can consider both factors simultaneously during the image generation process. The conditioning information provides specific semantic details such as the image category, style, or other relevant features, while the latent prior guides the model's randomness and diversity during the generation process.

However, it's important to note that the conditioning information can introduce additional biased information that is reflected in the generated images. The specific biases present in the conditioning information may affect the characteristics and attributes of the generated images. Therefore, when using diffusion-based conditional generation, it is crucial to consider and address any potential biases introduced by the conditioning information to ensure fair and unbiased results. In our work, we focus on optimizing the conditional generation process $pt$ of the text-guided diffusion model. By specifically targeting the $pt$ component, we aim to minimize the potential bias assumption in language embeddings. 

\section{Experimental Details}

\subsection{Keyword Dataset}
\label{dataset}
In our research, we selected keywords for fair image generation based on a thorough investigation detailed in \citep{diffusion-bias}. These chosen keywords cover a variety of job roles (refer to Table \ref{tab:occupations}) and emotional states (see Table \ref{tab:emotions}). Our experiments involve a total of 150 different occupations and 20 emotional states, ensuring a diverse and comprehensive range for a thorough examination of our proposed approach.
\begin{table}[h]
\centering
\renewcommand{\arraystretch}{1}
\vspace{-7pt}
\caption{Keywords about Emotions}
\begin{tabular}{lll}
\toprule
ambitious     & determined   & pleasant       \\
assertive     & emotional    & self-confident \\
committed     & gentle       & sensitive      \\
compassionate & honest       & stubborn       \\
confident     & intellectual & supportive     \\
considerate   & modest       & unreasonable   \\
decisive      & outspoken    & \\
\bottomrule
\end{tabular}
\label{tab:emotions}
\end{table}

\begin{table}[h]
\centering
\renewcommand{\arraystretch}{1}
\caption{Keywords about Occupations}
\begin{tabular}{lll}
\toprule
Accountant                & Facilities Manager            & Office Worker          \\
Aerospace Engineer         & Farmer                        & Painter                \\
Aide                       & Fast Food Worker              & Paralegal              \\
Air Conditioning Installer & File Clerk                    & Payroll Clerk          \\
Architect                  & Financial Advisor             & Pharmacist             \\
Artist                     & Financial Analyst             & Pharmacy Technician    \\
Author                     & Financial Manager             & Photographer           \\
Baker                      & Firefighter                   & Physical Therapist     \\
Bartender                  & Fitness Instructor            & Pilot                  \\
Bus Driver                 & Graphic Designer              & Plane Mechanic         \\
Butcher                    & Groundskeeper                 & Plumber                \\
Career Counselor           & Hairdresser                   & Police Officer         \\
Carpenter                  & Head Cook                     & Postal Worker          \\
Carpet Installer           & Health Technician             & Printing Press Operator \\
Cashier                    & Host                          & Producer               \\
CEO                        & Hostess                       & Psychologist           \\
Childcare Worker           & Industrial Engineer           & Public Relations Specialist \\
Civil Engineer             & Insurance Agent               & Purchasing Agent       \\
Claims Appraiser           & Interior Designer             & Radiologic Technician  \\
Cleaner                    & Interviewer                   & Real Estate Broker     \\
Clergy                     & Inventory Clerk               & Receptionist           \\
Clerk                      & IT Specialist                 & Repair Worker          \\
Coach                      & Jailer                        & Roofer                 \\
Community Manager          & Janitor                       & Sales Manager          \\
Compliance Officer         & Laboratory Technician         & Salesperson            \\
Computer Programmer        & Language Pathologist          & School Bus Driver      \\
Computer Support Specialist & Lawyer                        & Scientist              \\
Computer Systems Analyst   & Librarian                     & Security Guard         \\
Construction Worker        & Logistician                   & Sheet Metal Worker     \\
Cook                       & Machinery Mechanic            & Singer                 \\
Correctional Officer       & Machinist                     & Social Assistant       \\
Courier                    & Maid                          & Social Worker          \\
Credit Counselor           & Manager                       & Software Developer     \\
Customer Service Representative & Manicurist                & Stocker                \\
Data Entry Keyer           & Market Research Analyst       & Supervisor             \\
Dental Assistant           & Marketing Manager             & Taxi Driver            \\
Dental Hygienist           & Massage Therapist             & Teacher                \\
Dentist                    & Mechanic                      & Teaching Assistant     \\
Designer                   & Mechanical Engineer           & Teller                 \\
Detective                  & Medical Records Specialist    & Therapist              \\
Director                   & Mental Health Counselor       & Tractor Operator       \\
Dishwasher                 & Metal Worker                  & Truck Driver           \\
Dispatcher                 & Mover                         & Tutor                  \\
Doctor                     & Musician                      & Underwriter            \\
Drywall Installer          & Network Administrator         & Veterinarian           \\
Electrical Engineer        & Nurse                         & Waiter                 \\
Electrician                & Nursing Assistant             & Waitress               \\
Engineer                   & Nutritionist                  & Welder                 \\
Event Planner              & Occupational Therapist        & Wholesale Buyer        \\
Executive Assistant        & Office Clerk                  & Writer                 \\
\bottomrule
\end{tabular}
\label{tab:occupations}
\end{table}
\clearpage

\subsection{More Details on Evaluation Metrics}
\subsubsection{Human-CLIP}

\begin{figure}[h]
\centering
\includegraphics[width=0.8\textwidth]{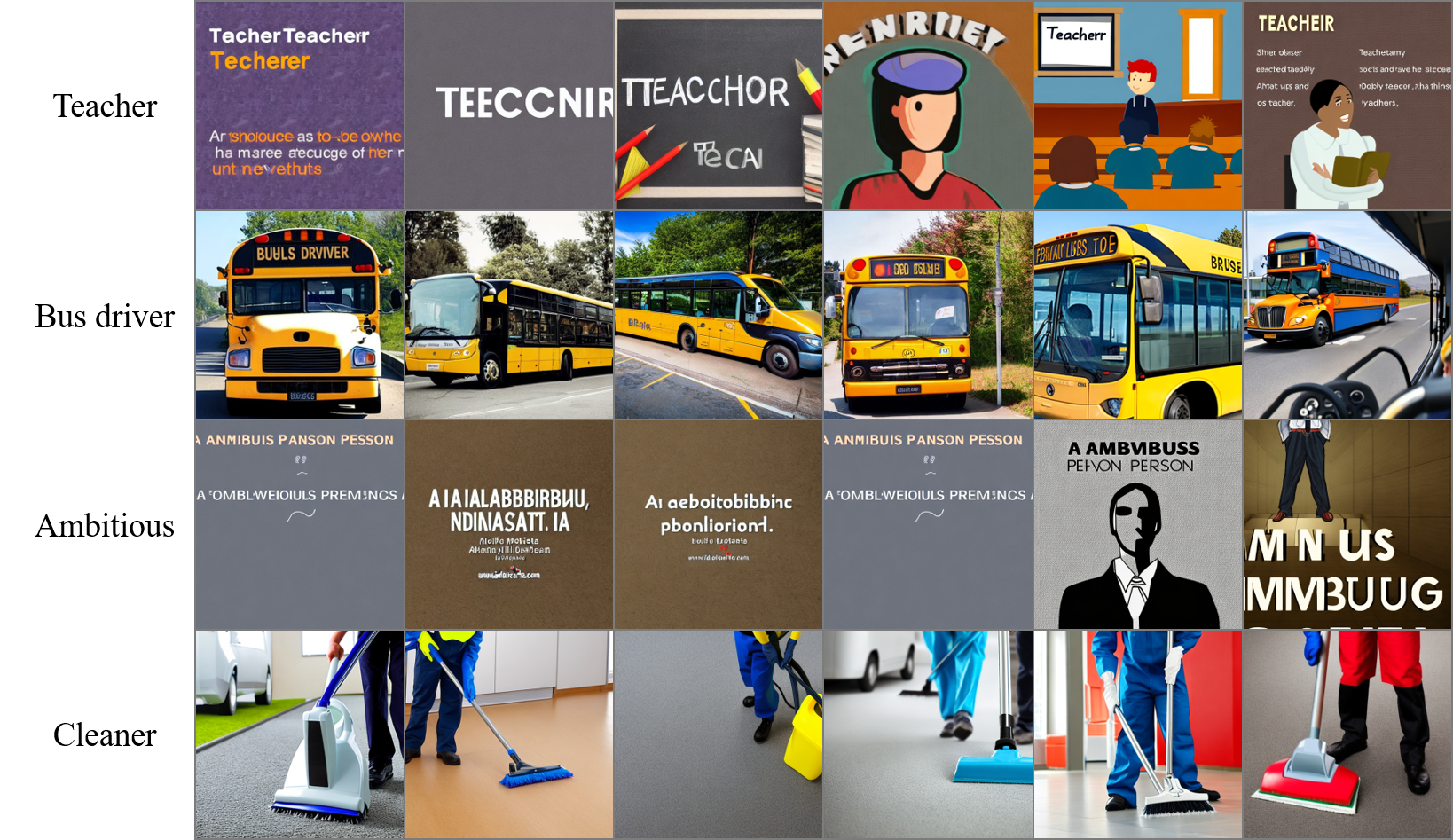}
\caption{Misalignment between generated images and human-related descriptions in Stable Diffusion.}
\label{fig:exp-humanclip}
\end{figure}

\label{human-clip}
The CLIP (Contrastive Language-Image Pretraining) Score serves as a prominent evaluation metric utilized for the assessment and comparison of semantic similarity between images and text in the context of generative models. This metric entails the utilization of pre-trained CLIP models, where images and text are inputted, and subsequent semantic relatedness is measured based on the similarity scores generated by the model. Higher scores indicate a greater degree of semantic relevance between the image and text, while lower scores suggest diminished semantic coherence. While CLIP demonstrates favorable performance across various tasks and domains, it is not exempt from limitations and challenges. Particularly in the case of text about human-related descriptions, the intricate and ambiguous nature of language poses difficulties for CLIP in achieving comprehensive understanding. Instances may arise where CLIP produces high scores for captions or input questions that exhibit inconsistency with the textual content, as depicted in the provided Figure \ref{fig:exp-humanclip}. 

Consequently, when confronted with human-authored descriptions, the stable diffusion may encounter constraints in its capacity to effectively generate images featuring human faces. We assess the efficacy of human face generating diffusion models, as presented in Table \ref{tb:human_pre}. In terms of occupations, the three different diffusion models, namely Stable Diffusion, Structure Diffusion, and Composable Diffusion achieved frequencies of 0.5914, 0.5641, and 0.6101, respectively, in generating facial images related to occupations. On the other hand, for emotions, the frequencies achieved by the Stable Diffusion, Structure Diffusion, and Composable Diffusion models were 0.6904, 0.5850, and 0.6274, respectively. Our approach demonstrates significant proficiency in generating realistic human facial features within generative images, exhibiting a notable improvement of over 30\% in performance when compared to alternative methods.

\begin{table}[!htbp]
\renewcommand{\arraystretch}{1.1}

\centering
\caption{Evaluation Results in Image Effectiveness for human frequency.\label{tb:human_pre}}
\begin{tabular}{ccc}
\toprule
Models & Occupation & Emotion \\
\midrule
Stable Diffusion     & 0.5914                 & 0.6904       \\
Structure Diffusion  & 0.5641                     & 0.5850     \\
Composable Diffusion & 0.6101                    & 0.6274          \\

\hline
    Fair Mapping-Gender          &  0.9229       &  \textbf{0.8950}          \\ 
Fair Mapping-Race          &    \textbf{0.9318}          & 0.7237          \\ 
\bottomrule

\end{tabular}
\end{table}

We propose the Human-CLIP metric as a remedy to the problem of not being able to capture all the subtleties in human-related textual generation based on the viewpoint presented in the literature \citep{metric-bias}. The Human-CLIP measure aims to accurately analyze the alignment between generated images and content that is relevant to humans by concentrating on analyzing the CLIP-Score related to human appearance, offering a way to gauge the effectiveness of text-to-image production and its applicability to humans by using the CLIP model's scores. It gets over the drawbacks of earlier metrics and offers academics and practitioners a new tool for evaluating and comparing the quantitative connection between generated images and human appearance.

Specifically, we selectively retain the CLIP-Score solely for generative images containing human subjects. For images lacking human descriptions, we assign a CLIP-Score value of 0:
$$
\text{Human-CLIP($img$,$t$)} = \left\{ 
\begin{aligned}
    CLIP(img,t),& \text{ if $img$ contains human}\\
    0, & \text{else}
\end{aligned},
\right.
$$
Where $CLIP(img,t)$ denotes the CLIP-Score between image $img$ and text $t$. To assess the overall performance of our model, we calculate the Human-CLIP score by averaging the Human-CLIP scores obtained for all generated images. Mathematically, the Human-CLIP score is defined as $\frac{1}{|I|} \sum_{{img,t} \in I} \text{Human-CLIP}(img,t)$, where $I$ is the set of image-text pairs $(img, t)$ in the generated image set. This calculation provides an aggregated measure of the model's ability to generate images that align with human perception and understanding, as evaluated through the CLIP model.

\subsubsection{Diversity}

\label{diversity}
\begin{figure}[h]
\centering
\includegraphics[width=0.8\textwidth]{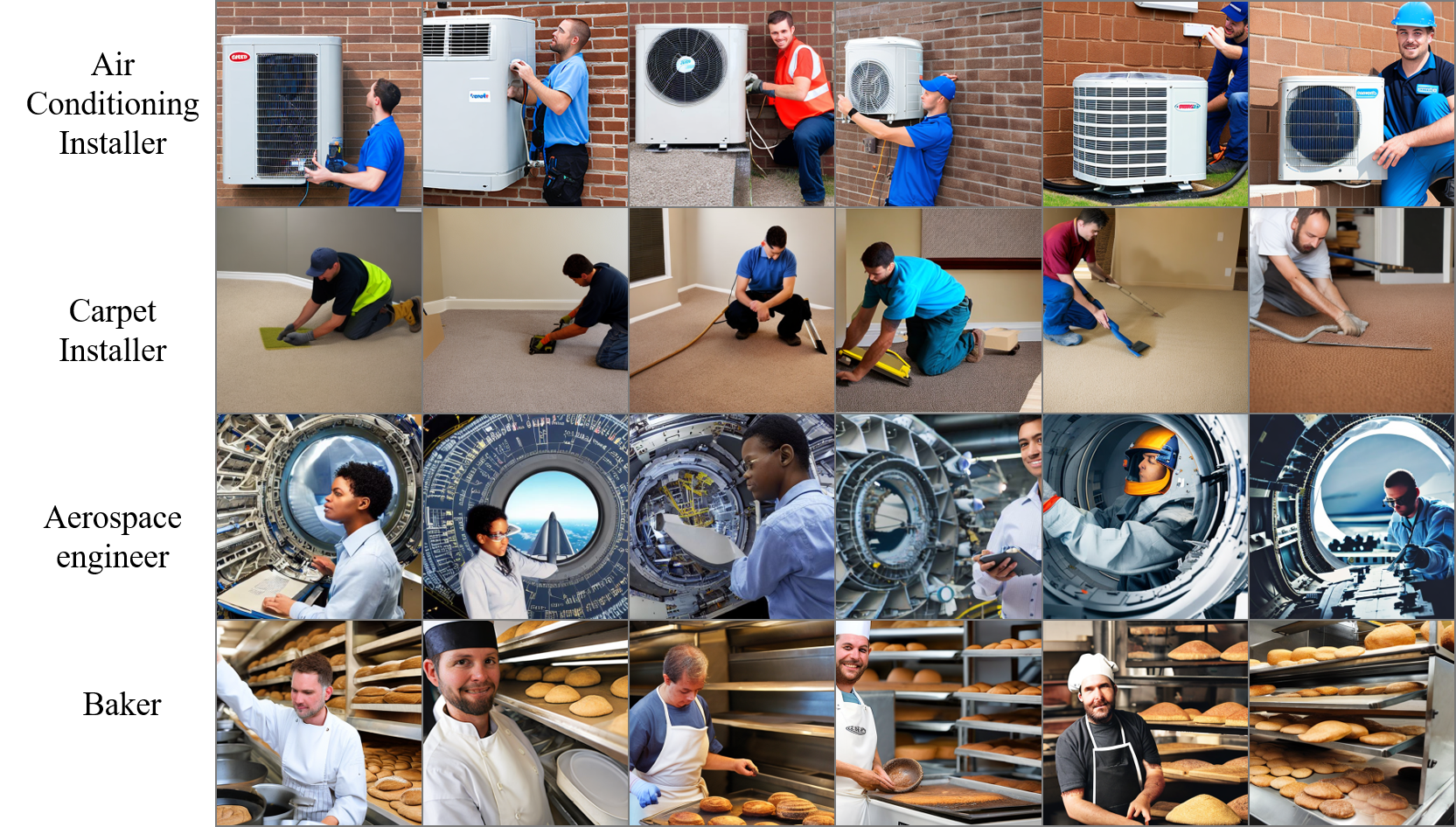}
\caption{characteristics of repetitive scenes in Stable Diffusion.}
\label{fig:exp-diversity}
\end{figure}

Stable Diffusion results show similarities to Mode Collapse in GANs, generating repetitive scenes with consistent backgrounds. As shown in Figure \ref{fig:exp-diversity}, for example, when generating images related to ``air conditioning installer'', the generated images often depict walls and air conditioning units with repetitive perspectives. We employ the intra-class average distance (ICAD) as a measure to assess the diversity of visual environments. To compute this metric, we focus on a particular class or category of generated images, such as ``Teacher'' or ``Pleasant''. By employing the squared Euclidean distance as the distance metric, we determine the average distance between all pairs of generated images within the chosen category: 
$$
ICAD(c) = \frac{1}{|S_c|}  \sum_{e_k \in S_c} \| e_k - \frac{1}{|S_c|} \sum_{e_i \in S_c} e_i \|_2,
$$
where $e_k$ and $e_i$ represents an individual generated image within category $c$, $S_c$ denote an image set generating controlled by $C$ .

Subsequently, we calculate the average value $\frac{1}{|D|} \sum_{c \in D} ICAD(c)$ across all keywords, where $D$ is a dataset that contains keywords. The average distance calculated within a keyword serves as a measure of dissimilarity or variability among the images belonging to that category. A smaller average distance indicates a higher degree of similarity or compactness among the images, suggesting a lower level of diversity within the visual environments they represent. Conversely, a larger average distance signifies greater variation among the images, indicating a broader range of visual environments captured by the generated images.

\section{More Experimental Results}

\subsection{Results on Human Preference}

\label{human}

\begin{figure}[h]
\centering
\includegraphics[width=\textwidth]{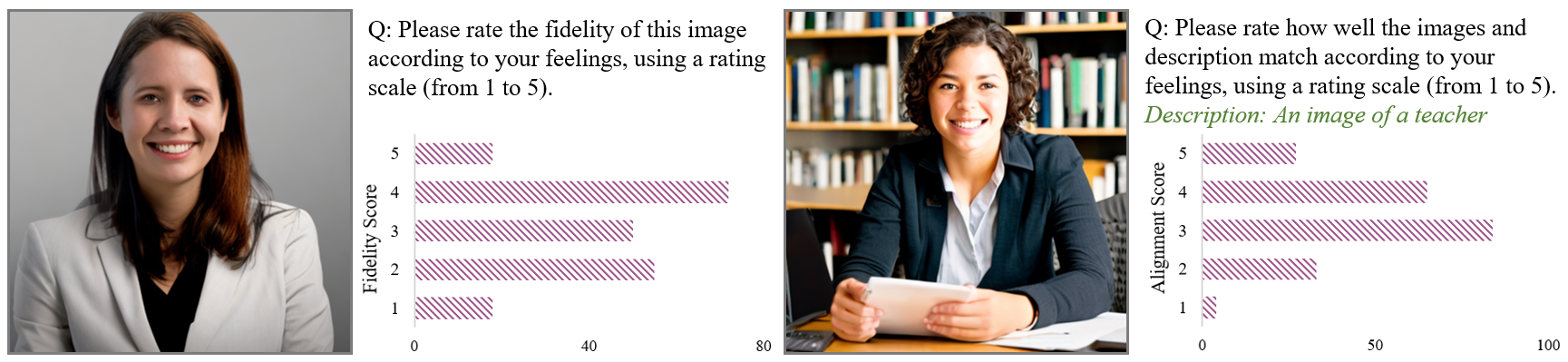}
\caption{The example for human preference in fidelity and alignment.}
\label{fig:exp-human}
\end{figure}

Figure \ref{fig:exp-human} showcases an example of the survey questions and the corresponding results. The evaluation of image authenticity and alignment with textual descriptions was conducted through a survey questionnaire. Participants over 200 from different academic backgrounds are presented with a series of generated images paired with corresponding textual descriptions. They are asked to rate the degree of fidelity in the images and alignment between the images and the provided descriptions from 1 to 5. To evaluate generated images, real images are included as a reference for comparison. For each task, we present users with two sets of 2 images along with the same input conditions for every method in each dataset. To mitigate potential biases stemming from preconceived notions of AI-generated images, we employ a data filtering process to exclude low-scoring samples. Subsequently, we compute the average performance of each method across diverse datasets to derive a comprehensive evaluation. Furthermore, we conducted a short survey to gather feedback on the reasons behind the scoring provided by the evaluators.

Specifically, we demonstrate our score criteria. In this study, we introduce our preference evaluation scale to assess the realism of images, using a scoring range from 1 to 5. The scale is as follows:
\begin{enumerate}
    \item[-] 1 Point: The image is likely AI-generated, displaying obvious artificial characteristics.
    \item[-] 2 Points: The image may be AI-generated with a realistic style, but discernible discrepancies from real imagery are present.
    \item[-] 3 Points: The image appears somewhat realistic but may have subtle flaws or unnatural features.
    \item[-] 4 Points: The image is very close to a real photograph, with details, colors, and lighting aligning with the real world.
    \item[-] 5 Points: The image is indistinguishable from real life, perfectly mirroring real-world standards in every aspect, including details, color, and lighting.
\end{enumerate}

Meanwhile, we employ a scale to evaluate the congruence between images and their corresponding captions, with a rating system ranging from 1 to 5:
\begin{enumerate}
    \item[-] 1 Point: Complete mismatch, the caption does not relate to the image.
    \item[-] 2 Points: Major discrepancy, the caption largely deviates from the image content.
    \item[-] 3 Points: Partial difference, there are noticeable mismatches between the caption and the image.
    \item[-] 4 Points: Minor discrepancy, the caption is almost in line with the image but with slight differences.
    \item[-] 5 Points: Perfect match, the caption accurately and completely describes the image.
\end{enumerate}

\section{More Visual Results}
\label{app-visual}
\begin{figure}[h]
\centering
\includegraphics[width=\textwidth]{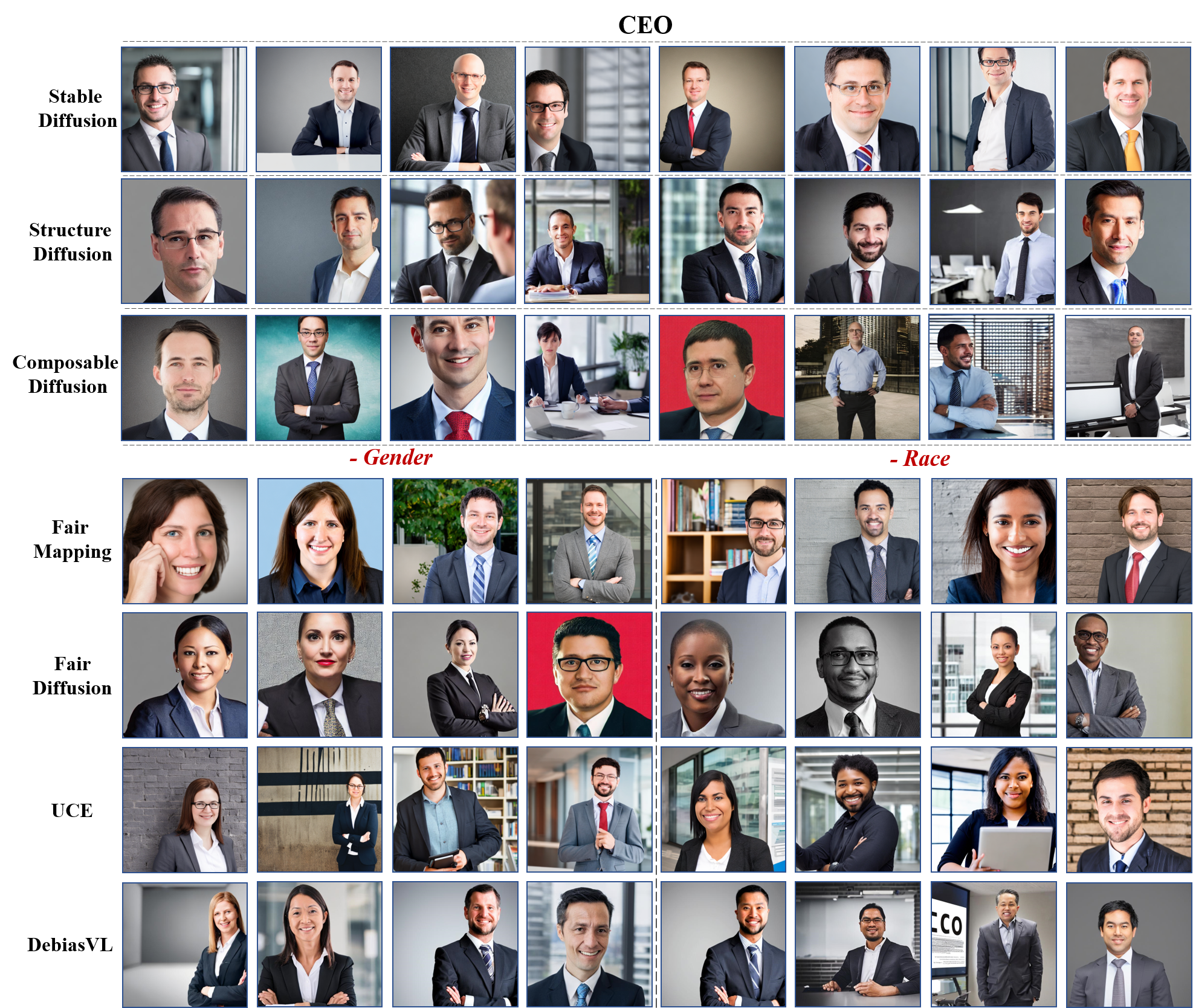} 
\caption{Comparison with different text-to-image methods: \textit{CEO}}
\label{fig:extra1}
\end{figure}

\begin{figure}[h]
\centering
\includegraphics[width=\textwidth]{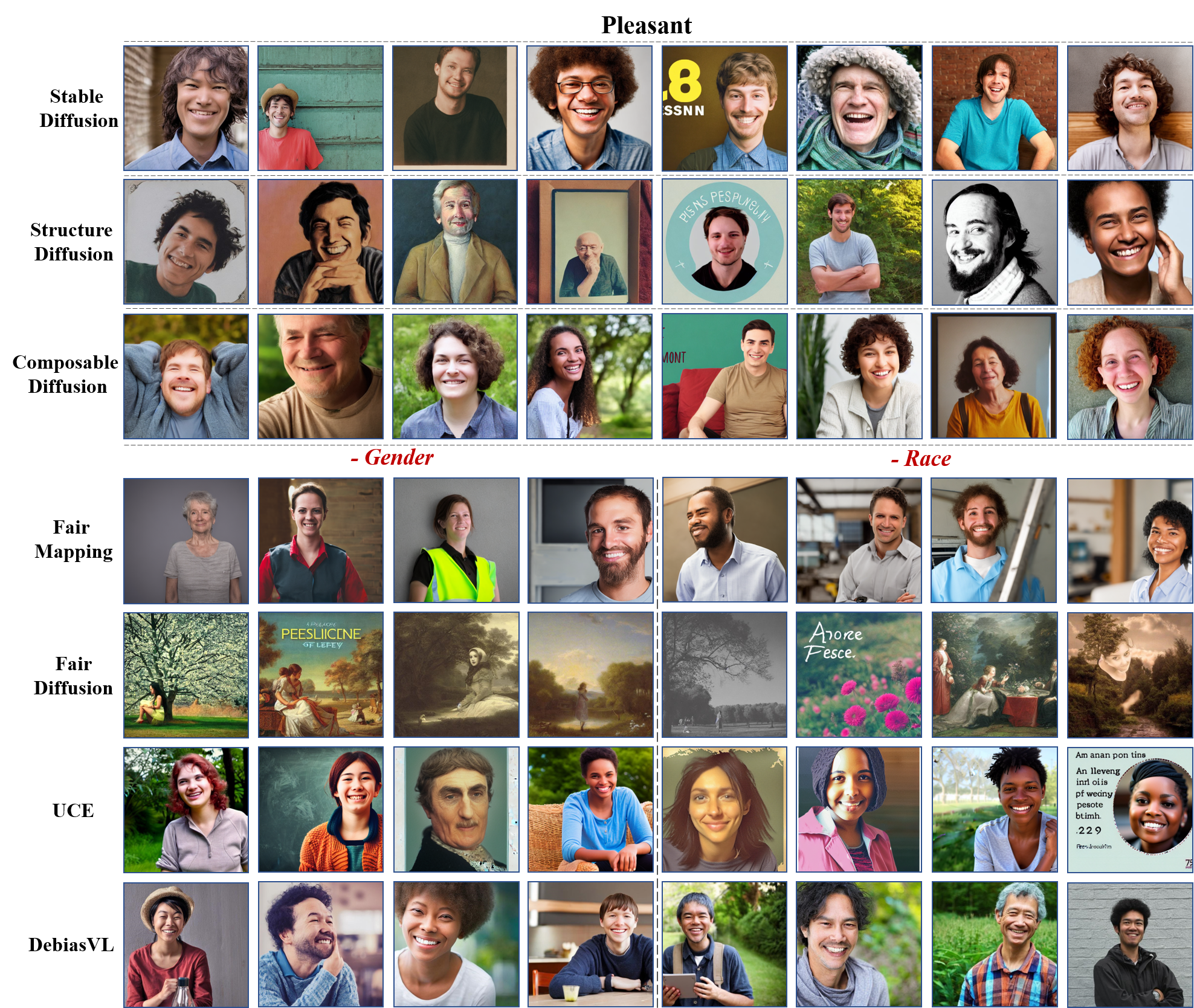} 
\caption{Comparison with different text-to-image methods: \textit{Pleasant}}
\label{fig:extra2}
\end{figure}

In the appendix, visual representations of the generated results are provided to further illustrate the research findings. These displays showcase samples of generated images corresponding to various categories or conditions. Figure \ref{fig:extra1} and Figure \ref{fig:extra2} show the images generated from keywords `CEO' and `Pleasant' respectively. Upon examination, it is noticeable that there is diversity in the generated images across different sensitive attributes. Specifically, Stable Diffusion tends to generate White male images with keywords `CEO' and `Pleasant'. In contrast, our method generates a greater number of female images and images depicting individuals with darker skin tones.

\end{document}